\RequirePackage{fix-cm}

\documentclass[twocolumn]{svjour3}          
\smartqed  
\usepackage{graphicx}
\usepackage{amsmath}
\usepackage{amsfonts,amssymb}
\usepackage{multirow}
\usepackage{cite}
\begin{document}

\title{Breast Tumor Classification and Segmentation using Convolutional Neural Networks
}

\subtitle{An efficient framework for breast cancer diagnosis based on CNNs and a combined fuzzy level-set approach}

\author{Parvin Yousefikamal \\
\email{parvin.yousefikamal@gmx.com} \\
Department of Electrical, Biomedical and Mechatronic Engineering, Qazvin Branch, Islamic Azad University, Qazvin, Iran
}

\authorrunning{Parvin Yousefikamal} 


\maketitle

\begin{abstract}
Breast cancer is considered as the most fatal type of cancer among women worldwide and it is crucially important to be diagnosed at its early stages. Throughout the past few years, a number of studies have been dedicated to the diagnosis of cancer from the medical images, yet no efficient method has been offered by these studies. In fact, the efficiency of the cancer diagnosis depends on two factors; first, the accuracy of the tumor area segmentation and calculation, and, second, the appropriateness of the features which are extracted from the images to classify the benignity or malignancy of a tumor. In the current study we aim to represent a fast and efficient framework which consists of two main parts:1- image classification, and 2- tumor region segmentation. \\
At the initial stage, the images are classified into the two categories of normal and abnormal. Since the Deep Neural Networks have performed successfully in machine vision task, we would employ the convolutional neural networks for the classification of images. In the second stage, the suggested framework is to diagnose and segment the tumor in the mammography images. First, the mammography images are pre-processed by removing noise and artifacts, and then, segment the image using the level-set algorithm based on the spatial fuzzy c-means clustering. The proper initialization and optimal configuration have strong effects on the performance of the level-set segmentation. Thus, in our suggested framework, we have improved the level-set algorithm by utilizing of the spatial fuzzy c-means clustering which ultimately results in a more precise segmentation. \\

In order to evaluate the proposed approach, we conducted experiments using the Mammographic Image Analysis (MIAS) dataset. The tests have shown that the convolutional neural networks could achieve high accuracy in classification of images. Moreover, the improved level-set segmentation method, along with the fuzzy c-means clustering, could perfectly do the segmentation on the tumor area. The suggested method has classified the images with the accuracy of 78\% and the AUC of 69\%, which, as compared to the previous methods, is 2\% more accurate and 6\% better AUC; and has been able to extract the tumor area in a more precise way. 

\end{abstract}

\section{Introduction}
\label{intro}
Breast cancer is an important and common type of cancer in the world, especially among women\cite{Azevedo15}. In a study by the American Cancer Society, in 2017, 252710 women were diagnosed with breast cancer, out of which 41070 were reported to die because of this type of cancer; whereas, out of 2470 men who were diagnosed with breast cancer, 460 died. Although we have witnessed a rise in the number of people who are diagnosed with breast cancer, the number of deaths is significantly decreased amongst all the ranges of age. This descending trend is attributable to the development of the medical imaging equipment and the early diagnosis of cancer which would bring about the treatment and controlling possibilities for this cancer. 

Microcalcifictions and masses are the two most important signs of the breast cancer and their automatic recognition is very important for prediction of cancer\cite{Mohamed14}.
Masses are divided into two groups: benign and malignat\cite{SUN15}. Benign mass is a non-dangerous lump whose early diagnosis can be quite effective in its comolete demolishment, whereas the malignant mass is a cancer that would potentially grow and expand to the other parts of body. Therefore, it is highly important to diagnose cancer it early stages for easier treatments and to save the patient from death.   

\textbf{Calcifications.}
Older women may occasionally find small calcium spots in their breasts. These spots are called “microcalsifications”, which, due to their tiny size, are not touchable. But, in mammography images, they would appear as small and bright spots. In most cases they are benign; yet, their observation in specific patterns may cause concern.  For instance, in some cases they grow in a form of a cluster or a line (they would have a cluster-like or linear growth), and this can be a sign of cancer\cite{Jiang14}. 

\textbf{Mass or Tumor.}
Breast are made up of a gland (Parenchyma) and a duct tissue; and the masses, which are described as occupying lumbs, may hide around this Parenchyma tissue of breasts; consequently, it may be somehow difficult to distinguish between the normal and abnormal area\cite{MassTum05}.  

Mammography has been introduced as the best method of cancer diagnosis in early stages\cite{Mammo18,Wang2014}. As the advantages of this method, one can point out to its low price and low risk for the patient. Although digital mammography is considered as the safest method for cancer diagnosis, its interpretation is a difficult task and for an inexperienced or tired radiologist the abnormalities of the mammography images may not be visible. For this reason, the application of the intelligent systems for the analysis of the images can prevent human errors in diagnosis. Studies show that the application of computer-aided diagnosis (CAD) systems as the second opinion systems can increase the sensitivity of the inexperienced radiologists from 62\% to 80\% and of the experienced ones from 77\% to 85\%\cite{PrePro2010,comCAD08}. 

The common dignosis systems operate in 3 different levels: preprocessing, feature extraction, and classification\cite{Mellisa15,Varsha15}. But these systems come with several challenages such as variation in form, size, borders and tumor tissue and also the existance of noise and extra objects in images.

During the past years, image processing methods have been applied successfully. One common method, for the recognition of abnormal area, is to use heuristics such as filtering or thresholding\cite{thresh05,Ogiela20,Gowri14}. However, these mehtods have some major drawbacks. To mention them, one can point out to their inapproperiate functionality when there is noise or other difficult imaging conditions. 
In fact, the statistical techniques have been taken into consideration for solving this problem. Brzakovic\cite{Brzakovic90} applied the fuzzy pyramid linking method to identify the area of the tumor in mammography images. Kegelmeyer\cite{Kegelmeyer1990} has also classified the pixels, using the vector computation of feature for each pixel and the binary decision tree. \\

The suggested method in the current study is a biphasic algorithm: in the first phase the mammography images are classified into normal and abnormal, using the Convolutional Neural Network (CNN), and in the second phase we are intended to reveal the tumor region in breast images with using the Level-set segmentation method based on the Spatial Fuzzy Clustering (LS-SFC). In order to classify the mammography images, first they need to be preprocessed; in better words, at this stage, the images are improved by a block-matching and 3D filtering (BM3D). Then, since the classification is done through a Convolutional Neural Network (CNN), there is a need for a large amount of data to feed the network. Therefore, it is necessary to improve the mammography images at the preprocessing stage. It is also worthy of note that Convolutional Neural Networks are capable of extracting features and, therefore, there is no need for manual design of the features extraction algorithms.  

In the second phase, to identify the tumor region in breast, we apply Level-set segmentation method based on the Spatial Fuzzy.

The rest of paper organized as follows:
In the next section we would review the previous literature in this field and would make a comparison among them. Then, in section \ref{sec_method}, we would elaborate our suggested method. Section \ref{sec_data} deals with the dataset and the preprocessing method, and finally in the last section we will discuss our experiments and conclusions. 

\section{Review of Literature:}

In this section, the previous works in the field will be briefly reviewed and the deep learning methods that have been applied in this respect will be introduced. 

\textbf{Features extraction}
A benign tumor usually has a circular and symmetrical form; whereas, the malignant tumor is asymmetrical and has a sharp tip. As the tumor shape features, we can mention area, perimeter, circularity, Fourier descriptor, compactness, eccentricity, etc. Generally, in comparison with the features of the shape, features of the tissue can reveal more information about the existing lumps\cite{feature14}. 

Ravishankar and Vishrutha\cite{Ravishankar15} used a combination of the wavelet features.
Torrents-Barrena\cite{Torrents14} et al. calculated the features of the tissue from the desired areas, using the multichannel gabor filter bank and, to calculate the features, they applied the multi-measurement processing windows in the tumor area. In order to diagnose the abnormalities in breast thermogram, V.Francis et al.\cite{Francis14} suggested the feature extraction method based on curvelet transform and they extracted the tissue features from the breast thermogram according to the curvelet area. To distinguish between benign and malignant tumors, Liu et al.\cite{Liu14} used the geometrical features and the tissue; this was mainly done to help them achieve a high level of accuracy in the segmentation of lumps.

\textbf{Classification} 
Setiawan et al.\cite{Sagiterry15} used the Artificial Neural Network (ANN) as a classification of the mammography images. The classification was finally done in two stages: first, data were classified into normal and abnormal, and then, the abnormal data were divided into benign and malignant. 
With several investigations, Hashemi et al.\cite{Hashemi15} introduced the Supporting Vector Machine (SVM) as the most successful method for the classification of the tumors in breast. Rodriguez-Lopez and Cruz-Barbosa\cite{Rodriguez-Lopez-10.1007/978-3-319-19264-2_28} compared the functionality of the Bayesian network models for determining the benignity and malignancy of the tumor. Bayesian networks are the models of probability, which would apply the technical knowledge.

\textbf{Application of deep learning in breast cancer diagnosis:}
As compared to other methods, deep-learning methods have performed efficiently in areas such as image recognition, image classification, and person identification\cite{DubrovinaCNN2018,CNN14,KIM-CNN-2017109}. The most important advantage of these methods is their auto feature identification, while, in classic methods features are manually designed. In 2012, Krizhevsky\cite{Krizhevsky12} introduced AlexNet with 5 layers of convolutional and 3 fully connected layers. This network successfully won the challenge of the image classification on ImageNet dataset. 

With the growing popularity of deep-learning methods, their application for medical purposes have also had effective performances; for instance, in the diagnosis of Alzheimer, cardio diseases, brain tumors, and etc. these methods have been found effective\cite{CNNAlzheimer17,CNNcardio16,CNNbrain17,CNNother17}. However, for breast cancer diagnosis and its region segmentation, these methods are not used a lot.

 In \cite{CNNmicroWang16,CNNmicro16}, authors used CNNs for representing features of microcalcifications.
Also, recently, Adaptive Deconvolutional Networks have been applied for this purpose\cite{AdaptiveDeconv17}. 

In 2015, Arevalo et al.\cite{Arevalo2015} obtained the ROC of 86\%, using a convolutional network as a feature extractor and a SVM as a classifier.

Jiao et al.\cite{JIAO16}, in 2016, could reach the precision of 96.7\% in the classification of the tumors into benign and malign. They used DDSM dataset and a convolutional network to serve them as a feature extractor and a SVM as a classifier. 

\textbf{Segmentation}
Segmentation in mammography images is defined as a stage where it is possible to locate the exact place of the lump in breast. 
To do a segmentation for Microcalcifications, Abdul Malek et al.\cite{Malek10} presented a combined method. In this suggested method, they applied the Region growing and Boundary segmentation.

R.S.C. Boos et al.\cite{RSCBoos12}, on the other hand, offered the Fuzzy C-mean clustering algorithm as a method for the segmentation of the desired area. In their suggested method, M.Rejusha and M.Kavith\cite{RejushaPectoral15}, first attempted to segment the pectoral muscle and then tried to diagnose the lump. While, Jumaat et al.\cite{Jumaatultra10} investigated the application of the active contouring methods for the segmentation of the masses in the ultrasound images of the breasts. 

X.Lin et al.\cite{XLinshape10} studied the mass segmentation ine respect to the Level set segmentation and the shape analysis. J.Liu et al.\cite{JLiu10}, also, offered an entirely automatic algorithm for the segmentation of the mammography images. In fact, what they suggested was to use the Marker-Controlled Watershed Algorithm and the level set method. 

\section*{Proposed Method}
\label{sec_method}
As shown in Figure \ref{method_arch}, the proposed method is a two-level algorithm. The first level is to classify the mammography images into two types of normal and abnormal and in the second level the existing tumor will be revealed in the images. 
In fact, what is being done in the first level is to classify the mammography images into cancerous and noncancerous, using the Convolutional Neural Network (CNN). Due to having convolutional layers and, also, kernels, CNN model is capable of extracting the features from images automatically; therefore, the need for manual design of the algorithms is eliminated. The next level of our suggested method is to determine the area of the tumor which, in this study, is done through Level-set area segmentation based on the Spatial Fuzzy Clustering (LS-SFC)\cite{LSSFC15}. In other words, the segmentation is primarily done by utilizing the Spatial Fuzzy Clustering, which contains spatial and local intensity information and then the results are used to improve the level-set segmentation. 

\begin{figure*}
\centering
\includegraphics[width=0.90\textwidth]{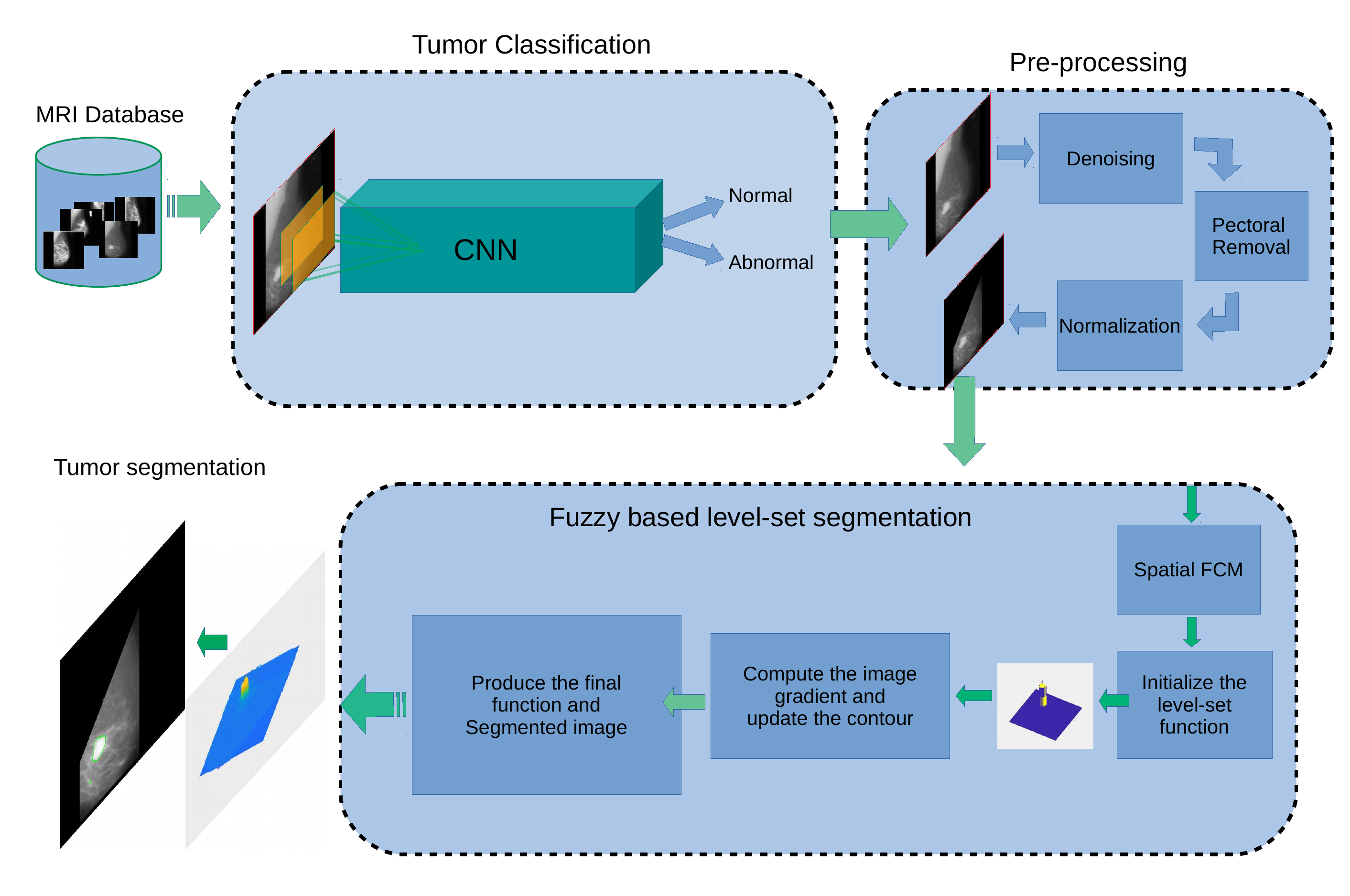}
\caption{Proposed framework for breast tumor classification and segmentation. In this framework, we train a Convolutional network to classify input images into normal or abnormal categories. In the next step, we utilize image enhancement techniques to improve images and remove pectoral muscles or unrelated objects. In the final stage, we employ a spatial fuzzy based level-set approach to segment the tumor region of abnormal input image.  }
\label{method_arch}
\end{figure*}

\subsection{Classification based on CNN approach}

\begin{figure*}
\centering
\includegraphics[width=0.8\textwidth]{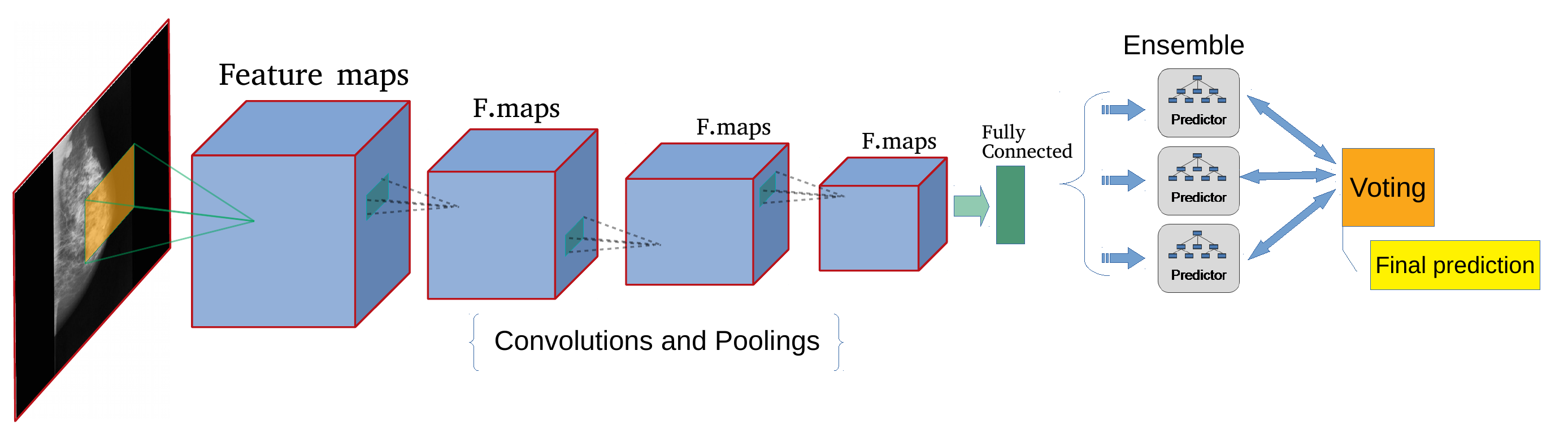}
\caption{Architecture of our CNN network. The architecture includes convolutions and poolings to extract deeper features for tumor classification. In the last stage, we utilize an ensemble to get final prediction by majority voting method.} 
\label{method_arch}
\end{figure*}

In CNN, convolutional layers are responsible for the features extraction. In fact, the convolutional layer is the main core of the CNN architecture. The first convolutional layer of this network extract features of the surface, such as edges, contours, and corners and outer layers extract the semantic features from the images.

The max pooling layer is located between the convolutional layers and its major responsibilities are to decrease the size of the feature map, to become resistant against noise and distortion, and to increase the speed of convergence. 

In the convolutional Neural Network (CNN), we utilize a soft-max function to obtain the class label. The soft-max function calculates the probable outcome of each category, using the below equation:
\begin{equation}
p(y=i|x, w_{1},...,w_{M}, b_1,...,b_M)= \frac{e^{w_ix+b_i}}{\sum_{j=1}^Me^{w_jx+b_j}} 
\end{equation}

\begin{equation}
\widehat{y}=\operatorname*{arg\,max}_{i} p(y=i|x, w_{1},...,w_{M}, b_1,...,b_M)
\end{equation}

Prior to the segmentation of the tumor area in breast images, first, the pre-processing performed in order to remove noise and the extra (unnecessary) information, which will be briefly explained in section \ref{sec:pre_process}. 

 \subsection{Segmentation}
 In the proposed method, the level-set based on Spatial Fuzzy Clustering algorithm for segmentation of tumor regions is applied. In fact, breast tumors are difficult to be diagnosed due to their complex from and figure; therefore, since the level-set method uses the dynamic variable boarders to do the area segmentation, it can be considered as a suitable method to find the tumors. However, due to its large amount of calculations, this method is not ideally applicable and it would require manual settings for initialization and other controlling parameters. To facilitate the segmentation task, we suggested a combination of the Level-set algorithm and the Spatial Fuzzy Clustering algorithm. To improve performance of level-set algorithm, we use the Spatial Fuzzy Clustering method to initialize the function. We will elaborate on this algorithm in the following section. 

\subsubsection{Spatial Fuzzy Clustering}
C-Means Fuzzy algorithm is also defined by the membership function and it attempts to minimalize the following target function:
\begin{equation}
J = \sum_{n=1}^N\sum_{m=1}^C\mu_{mn}^L||i_n-v_m||^2
\end{equation}
In which $\mu_{mn}$ is the membership function (the degree to which the data belong each cluster), $L$ is the Fuzzification parameter, $v_m$ center of $m^th$ cluster,$i_n$ is the image pixel, $N$ is the labeled objects (exp. The number of pixels in the image $N=N_x\times N_y$), $C$ is the number of clusters and “.” Indicates the Euclidean distance. 

The membership functions should also satisfy the following conditions:

\begin{equation}
\sum_{m=1}^{C}\mu_{mn}=1~;~ 0\leqslant \mu_{mn} \leqslant 1 ~;~\sum_{n=1}^{N}\mu_{mn}>0
\end{equation}

The above-mentioned equation illustrates that the memberships can’t go minus and the sum of membership coefficients for one element on the clusters equals 1. The target function is also frequently updated by the $\mu_{mn}$ and $v_i$ :
\begin{equation}
\mu_{mn}=\frac{||i_n-v_m||^{\frac{-2}{L-1}}}{\sum_{k=1}^{C}||i_n-v_k||^{\frac{-2}{L-1}}}
\end{equation}

\begin{equation}
v_i=\frac{\sum_{n=1}^{N}\mu_{mn}^Li_n}{\sum_{n=1}^{N}\mu_{mn}^L}
\end{equation}

Although the FCM clustering algorithm offers some positive points, such as its being able to operate without supervision and its permanent convergence, it also has some drawbacks. Among its drawbacks we can mention its sensitivity to noise. FCM algorithm lacks the spatial information and this would cause sensitivity to noise and artifacts. Recently, several researchers have made lots of attempts in this field to improve the segmentation performance. Sudip Kumar Adhikar et al.\cite{SudipSPFCM15} have presented a Spatial Fuzzy Clustering algorithm (SpFCM) to perform the area segmentation in brain MRI images. This algorithm solve the problem of noise sensitivity and reduces the non-homogeneity in the images. They introduced a function of possibility which would demonstrate the possibility of neighboring pixels to be belonging to the cluster. Then, they presented a new membership function, using the spatial information. Finally, according to the new local and global membership function, they provided new clustering centers and a weighted membership function. In their algorithm problem of noise sensitivity solved and degree of non-homogeneity in MRI images is decreased and, in general improved the performance of the segmentation task. 

\subsubsection{Level-set based on Spatial Fuzzy Clustering}
One of the most popular and efficient methods of area segmentation in medical images, is the active contouring model which was introduced by Kass\cite{Kass1988}. This Model is also known as the “Snakes” model, which is based on the development of the shaped curve; simply put, it is actually a dynamic curve. 

Since the Level-set method applies variable dynamic boarders to do the area segmentation, it is considered an appropriate method; however, due the large amount calculations it requires, this method decreases the efficiency. Therefore, what we have suggested to solve this problem is to combine the Level-set method with the Spatial Fuzzy Clustering. 

The new algorithm of the Level-set starts with Spatial Fuzzy Clustering and the results of the Spatial FCM are used for the initialization of the of the level-set algorithm. In other words, the new algorithm is automatically responsible for the initialization and the parameter assignment of the level-set segmentation. In fact, having considered the spatial information, the fuzzy clustering method would determine the approximate contours of the desired items in the mammography images. The level-set function is initialized through the below equation:

\begin{equation}
\phi=-4\varepsilon (0.5-B_k)
\label{equ_26}
\end{equation}
In which $B_k$ is the binary image and is calculated as below:
\begin{equation}
B_k=R_k\geqslant b_0
\label{equ_27}
\end{equation}
In the above equation $b_0\in (0,1)$ is an adaptable threshold. Also, in equation \ref{eque_28}, $\varepsilon$ is a regulator for Dirac function:

\begin{equation}
\delta_\varepsilon (x)=\left\{\begin{matrix}
0, ~&~ |x|>\varepsilon \\ 
\frac{1}{2\varepsilon }\left [ 1+ cos ( \frac{\pi x}{\varepsilon } ) \right ],~ &~ |x|\leqslant \varepsilon 
\end{matrix}\right.
\label{eque_28}
\end{equation}

In the equation~\ref{equ_27}, $R_k$ is the resulted image of level-set approach using spatial fuzzy clustering. Therefore, level-set functions starts from a binary region based on equation~\ref{equ_26}:

\begin{equation}
\phi(x,y)=\left\{\begin{matrix}
C,~ &~\phi _0(x,y)<0 \\ 
 -C,~&~ otherwise 
\end{matrix}\right.
\end{equation}
And finally, evolution of level-set would be as following:
\begin{equation}
\phi ^{k+1}(x,y)=\phi ^k(x,y)+\tau\left [\mu\xi (\phi )+\xi (g,\phi ^k)  \right ] 
\end{equation}

\section{Dataset}
\label{sec_data}
\begin{figure}
\centering
\includegraphics[width=0.49\textwidth]{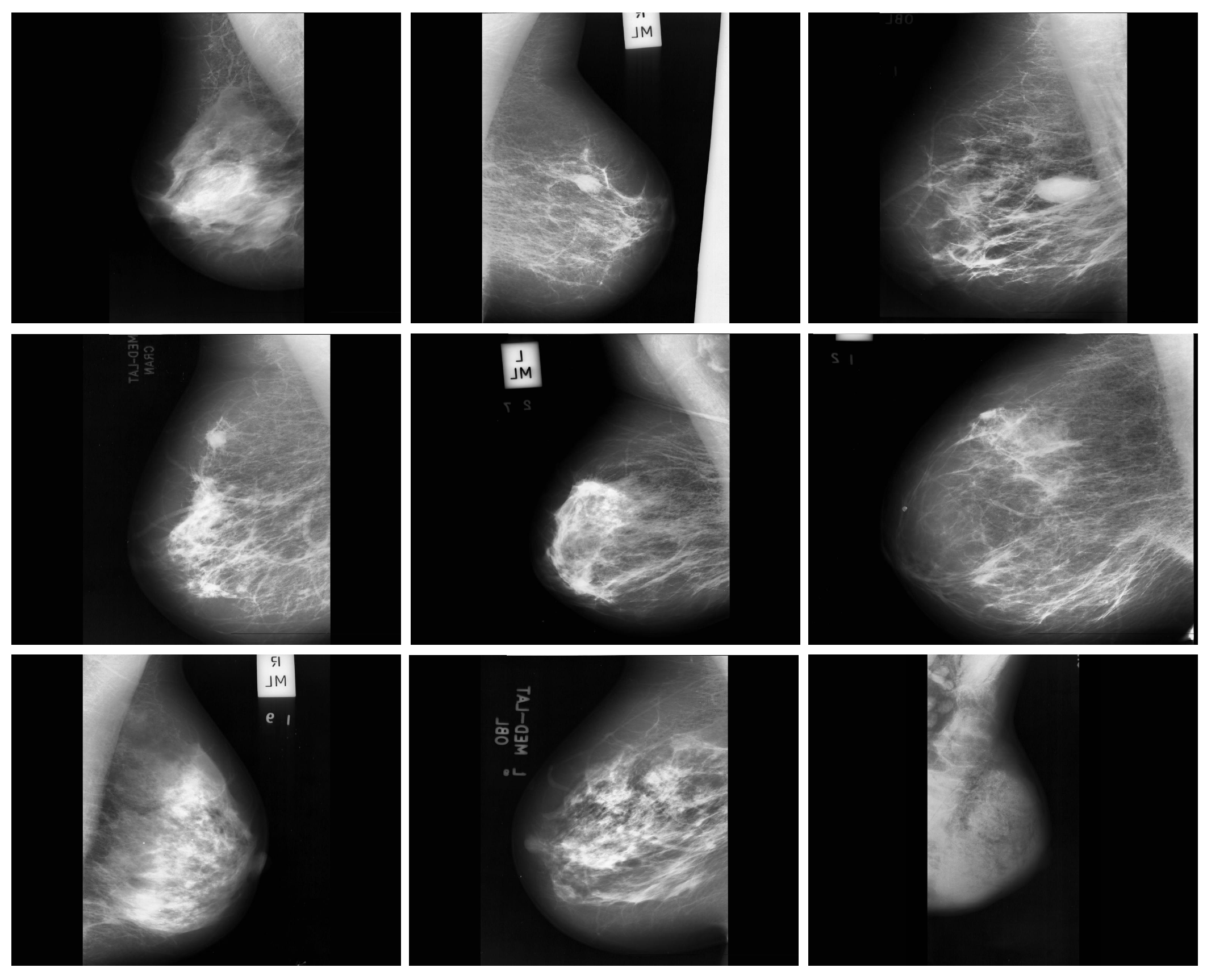}
\caption{Examples from the MIAS dataset. In this dataset, the pectoral muscle and out-layer objects make the segmentation a challenging task. Therefore, in this paper we propose several techniques to improve images before segmentation. }
\label{db_fig}
\end{figure}

In the present study, we conducted our experiments using the images from the MIAS dataset~\cite{DatasetMIAS15}. This dataset contains 322 mammography images (119 of which are abnormal images and the rest are normal). All mammograms are taken in a Medio-lateral oblique, with the dimension of 1024*1024 a resolution of 200. Several samples of this dataset are shown in Fig. \ref{db_fig}. 

The radius of the smallest mass in this database is 3 m pixel and the one for the biggest mass equals 197 pixel. This dataset is also divided into 7 different categories.  

\section{Pre-processing}
\label{sec:pre_process}
It's quite difficult for physicians to interpret mammography images due to the challenges such as noise, low contrast and variations.
 In fact, it is seriously important to eliminate the noise in order to analyze the medical images. Therefore, with pre-processing we try to improve images and make breast structure more reliable.

In order to improve the mammography image quality Sundaram et al.\cite{SundaramAdaptMed15} have suggested the adaptive median filter which is able to eliminate the impulse noise. Görgel et al.\cite{Gorgel13}, on the other hand, have applied the homomorphic filtering in so doing. This filter uses the lightening-reflection model to improve the compression of the lightening area and to make better contrast. 

Lashari et al.\cite{Lashariwavelet15} discussed the noise elimination from the mammography images through the wavelet filters. 

Prabha s et al.\cite{Prabha14} applied the block matching and 3D filtering algorithm in order to improve the thermogram images of breast, which is a powerful model, and this method significantly eliminated the noise from the images. Elahi et al.\cite{Elahi14}, also, used the same algorithm (BM3D) to reduce noise in brain Magnetic Resonance Image (MRI). In fact, they enhanced the performance of BM3D through the denoising method.

\begin{figure*}
\centering
\includegraphics[width=0.97\textwidth]{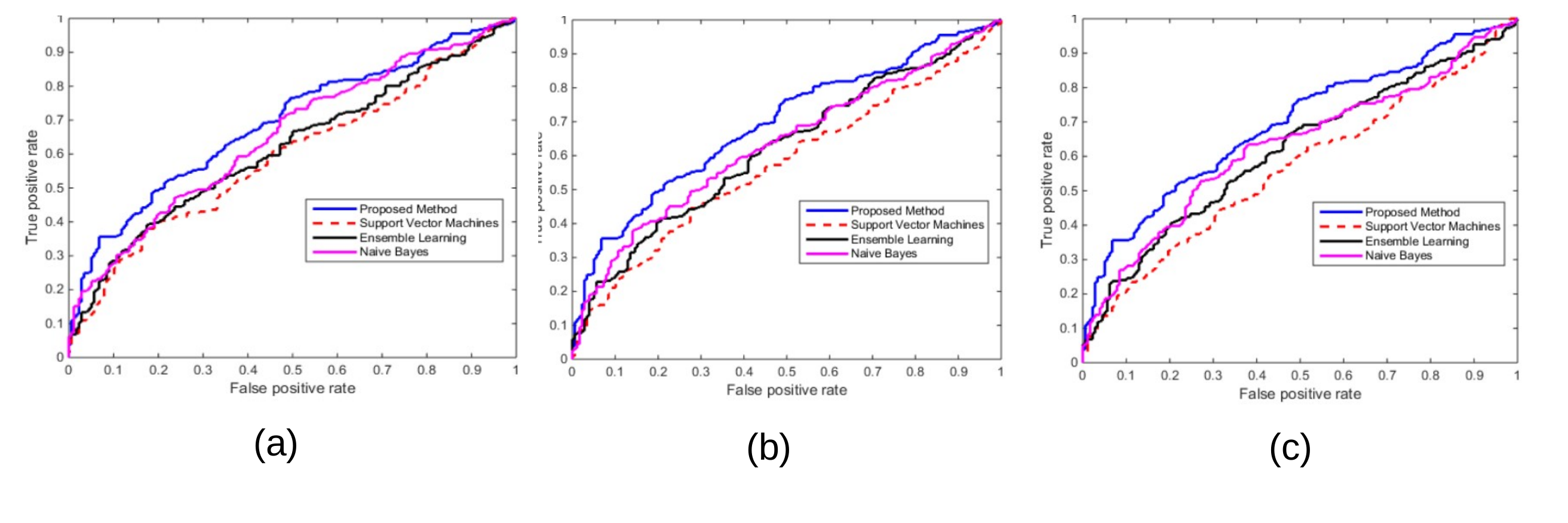}
\caption{ROC curve of different approaches for cancer recognition. a) with 100 features, b) with 200 featues and c) with 300 features.}
\label{roc_fig}

\end{figure*}   
    


\subsection{Denoising based on BM3D algorithm}
To eliminate noise we have applied the BM3D method, whose parameters are determined as below (table 2-5): 

The parameters are defined as:
\begin{itemize}
\item $N^{hard},~N^{wie}$: The maximum number of pieces which are similar to the reference block (located in the 3D group)
\item $\lambda^{hard}_{3D}$: The applied coefficient for hard thresholding
\item $\tau^{hard}$: Hard threshold
\item $\tau^{wie}$: Wiener threshold 
\item $k^{hard},~k^{wie}$: The size of the reference  block 
\end{itemize}
    
    The parameter $\lambda^{hard}_{3D}$ defines the thresholding surface of the 3D group in the transformation area and is usually given the value of 2.7. The thresholds $\tau^{hard}$ and $\tau^{wie}$ are  dependent to the value of $\sigma$. In case the value of noise is high, $\tau^{wie}$ and $\tau^{hard}$ are 3500 and 5000 respectively, whereas, if this value is low these amounts will be 2500 and 400 respectively. For a low $\sigma$, the block is of a smaller size and its details are preserved. For a high $\sigma$, on the other hand, a bigger block size is preferred, since more details are destroyed because of the noise. 
In this method, the 3D transformation is also done through the 1D and 2D linear transformation. 

\subsection{Data Augmentation}

We utilize data augmentation techniques to increase the number of samples in dataset to improve network training process. We do this after the noise reduction. Data augmentation also helps to reduce overfitting problem.
The following transformations are implemented:  

\textbf{Rotating:} the rotation is done with the interval of 0,360 from the angles. The mean-pixel value of the training set fills up the white corners which are the results of the rotation. 

\textbf{Mirroring and scaling:} in this method images are randomly made bigger or smaller; and also randomly the images are mirrored. 

\textbf{Cropping:} the size of the images is changed to 224 pixels and this is done along their shorter side, meaning that the size other side is accordingly changed, and 224*224 sized images are randomly made from the resized image. 

\textbf{Image shifting:} in this method, images are shifted and cropped in horizontal and vertical direction to change into the desired size and the amount of shifting is randomly selected.

4 random rotations and 4 random crops are done for each rotation in every image in the training set so that the size of the set is 16 times bigger. To make larger training dataset random mirroring and shifting of input is also done.

\begin{table*}

	\begin{center}
		\begin{tabular}{c|c|cc|cccccc|}
        \hline
			$\#$ Features& Method & ~AUC ~ & ~g-mean ~ & ~F-measure~ & Recall~ & ~Precision~& ~Specificity~& ~Sensitivity~& ~AC~\\
			\hline
        \multirow{4}{*}{\rotatebox{45}{100}}   &Ensemble       & 0.62          & 0.72  & 0.81  & 0.78 & 0.85 & 0.66 & 0.7836 & 0.75 \\
                                       &SVM     & 0.59          & 0.73   & 0.82  & 0.78 & 0.87 & 0.69 & 0.7813 & 0.75 \\
                                           & Naive Bayes      & 0.65          & 0.74   & 0.82  & 0.76 & 0.90 & 0.71 & 0.76 & 0.75  \\
                                               \hline
                                                
   \multirow{4}{*}{\rotatebox{45}{200}}        &Ensemble       & 0.62          & 0.70  & 0.81  & 0.77 & 0.84 & 0.64 & 0.77 & 0.74 \\
                                       & SVM      & 0.57         & 0.74   & 0.83  & \textbf{0.79} & 0.87 & 0.69 & 0.79 & 0.76 \\
                                           & Naive Bayes      & 0.63          & 0.74   & 0.83  & 0.77 & 0.90 & 0.71 & 0.77 & 0.75  \\
                                                \hline

   \multirow{4}{*}{\rotatebox{45}{300}}  & Ensemble       & 0.62         & 0.71  & 0.81  & 0.77 & 0.85 & 0.65 & 0.77 & 0.74 \\
                                       & SVM      & 0.56          & 0.73   & 0.82  & 0.78 & 0.87& 0.68 & \textbf{0.79} & 0.76 \\
                                           & Naive Bayes    & 0.63         & 0.75   & 0.83  & 0.78 & 0.89 & 0.71 & 0.78 & 0.76  \\ \hline
                                               & CNN     &\textbf{0.69}         & \textbf{0.78}   & \textbf{0.85}  & 0.78 & \textbf{0.93} & \textbf{0.78} & 0.78 & \textbf{0.78}  \\ \hline
            \hline
            
		\end{tabular}
	\end{center}
	\caption{Comparsion of proposed method with the state-of-the-art with different number of features. }
	\label{table_2}
	\vspace{-1mm}
\end{table*}

\subsection{Image enhancement}

\begin{figure}
\centering
\includegraphics[width=0.49\textwidth]{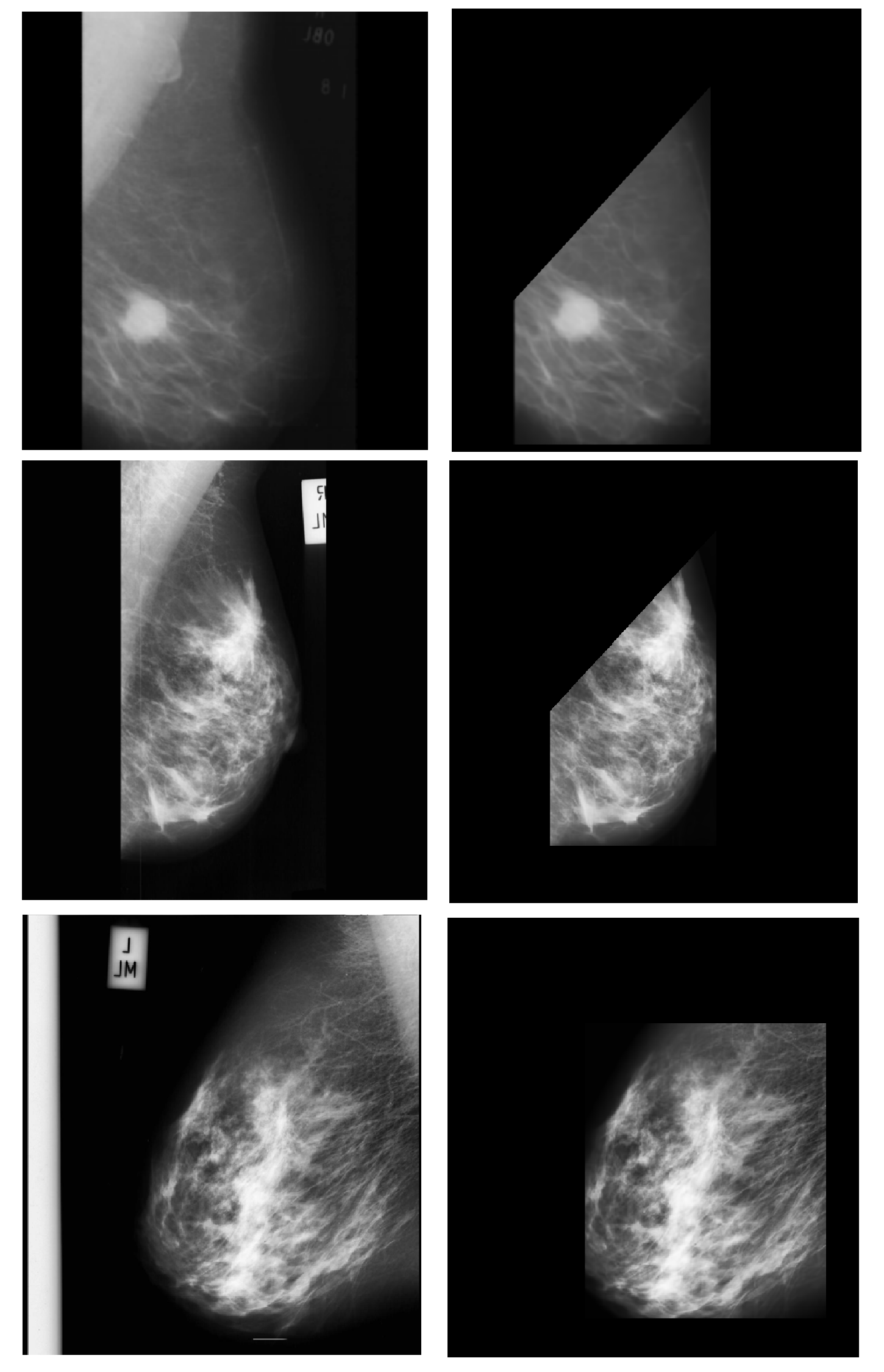}
\caption{Enhanced images after denoising and pectoral removal. First Column shows the input MRI images and second column shows images after enhancement.}
\label{preprocess_fig}
\end{figure}

Below you can read about the steps we’ve taken in our suggested method to achieve an accurate image which is free from undesirable factors. See Fig. \ref{preprocess_fig} \\  
\textbf{Median filtering:} using the median filtering to eliminate the high frequency components. \\
\textbf{Normalization:} images are taken through a variety of methods; consequently, a general demonstration of breast might not be possible and, as a result, normalization is being applied. All mammograms are also transferred to the fixed intensity range between $r_1$ and $r_2$, $0\leqslant r_1<r_2 \leqslant 255$. Assuming that the image $g_i(x,y)$ has the maximum grey level of $max~G_i$ and minimum grey level of $min~G_i$, normalization is done as below: 
\begin{equation}
g_k(x,y)=r_1+(g_i(x,y)-min~G_i)\times(r_2-r_1)/(max~G_i-min~G_i)
\end{equation}
\textbf{Thresholding:} using the thresholding technique, the mammography images of the grey level are transformed into binary images. \\

\textbf{Removal of the pectoral muscle}:
Pectoral has an approximate triangular shape which is located above, on the left, or on the right side of the image. The structure of this muscle is so much similar to that of the neighboring area of the tumor. Therefore, its presence of this muscle in mammography image will cause disruption in tumor diagnosis and may also result in a false positive diagnosis. Thus, detecting the pectoral muscle and its removal is of high importance in the correct diagnosis of cancerous areas\cite{PectoralCardo10,PectoralMajeed13}. 
We have applied a similar method for the removal of this area.

\section{Results}

\subsection{Evaluation measures}
To evaluate the suggested method different factors are being applied. The ROC diagram illustrates true positive rate (TPR) to the false positive rate (FPR); The value of TPR is located on axis Y and the one of FPR is on axis X. Also, the area under the curve ROC (AUC) shows to what extend this test is ideal. Evaluation parameters: \\
\textbf{Accuracy:} Number of cases with or without cancer that are classified correctly.  
\begin{equation}
AC=\frac{TP+TN}{TP+FP+TN+FN}
\end{equation}
\textbf{Sensitivity:} The proportion of cases that are correctly diagnosed with cancer.
\begin{equation}
SE=\frac{TP}{TP+FN}
\end{equation}
\textbf{Specificity:}  the proportion of negative cases that are correctly recognized.
\begin{equation}
SP=\frac{TN}{TN+FP}
\end{equation}
\textbf{Precision:} the proportion of detected cases that are actually cancerous.
\begin{equation}
PR=\frac{TP}{TP+FP}
\end{equation}
\textbf{Recall:} this factor demonstrates the number of true positive cases to the related cases.
\begin{equation}
RE=\frac{TP}{TP+FN}
\end{equation}
\textbf{F-measure:} is a factor which can show the performance of the classification with a single number.
\begin{equation}
F-measure=2\times \frac{RE\times PR}{RE+PR}
\end{equation}
\textbf{Geometrical mean:} when the number of true positive and true negative cases are at the highest level and the difference between the two group of cases is at its lowest level, this factor will be at the highest level. 
\begin{equation}
\sqrt[]{\frac{TP}{TP+FN} \times \frac{TN}{TN+FP} }
\end{equation}

\subsection{Classification Results}

For training the CNN network, we utilize early layers of AlexNet\cite{AlexNet12} for our architecture. We also make use of batch normalization \cite{Normalize15} for better training. The network receives the input image with size of $256\times 256$. The final layer is a soft-max layer with 2 output for classifying inputs into two classes (with cancer or without cancer). In addition, the learning rate equals with 0.001 and epoch is 20 in our setting.

\begin{figure*}
\centering
\includegraphics[width=0.99\textwidth]{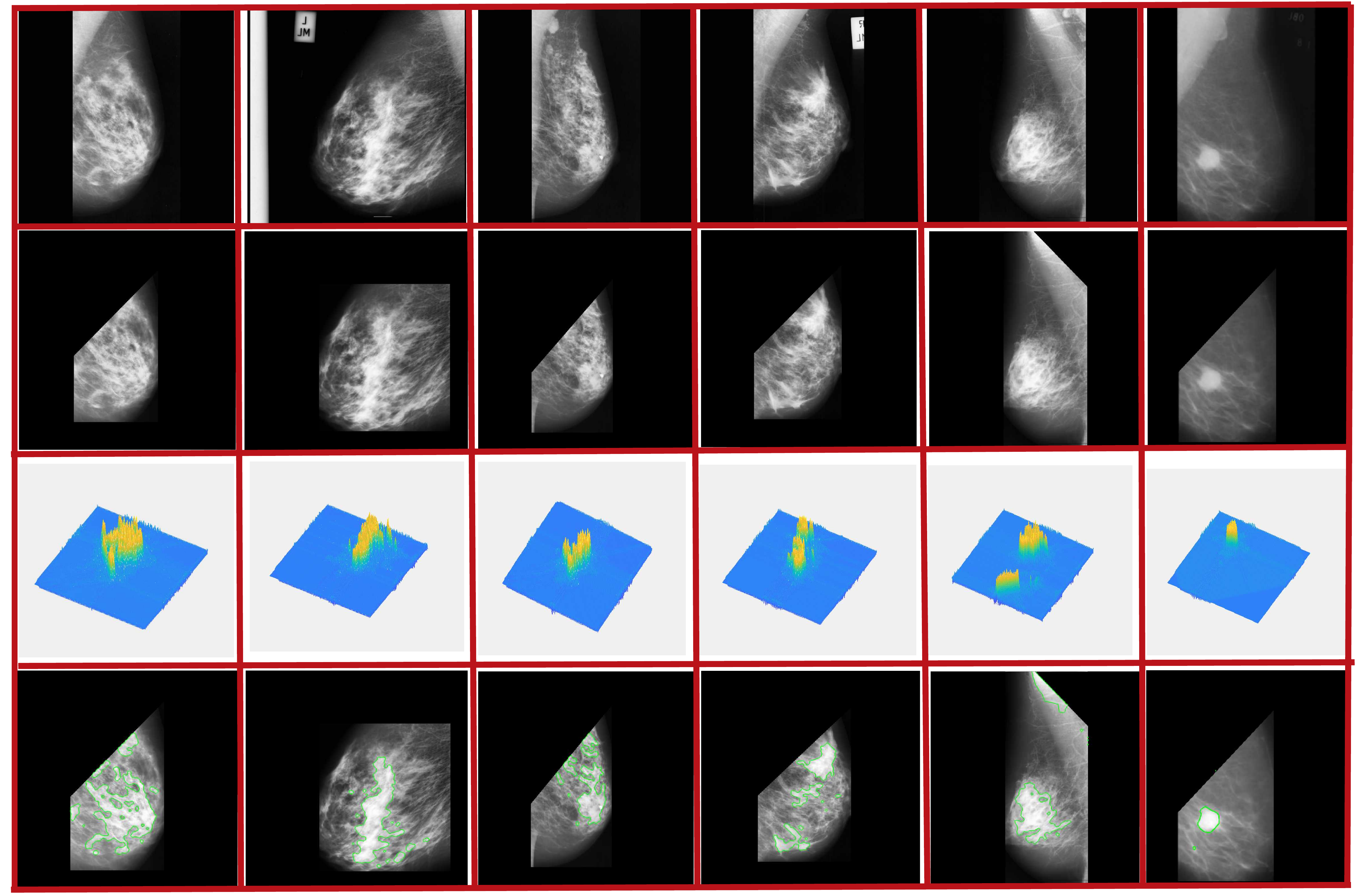}
\caption{Qualitative result of proposed method. First row shows the input images, second row shows the pre-processed images, third row is the final level-set functions and last row shows the final segmented tumor. }
\label{qualitative_results}
\end{figure*}


As it can be seen in Table\ref{table_1}, the algorithm of the current study could successfully detect 0.7786 sick people, 0.7876 healthy people, and 0.9288 people who actually have cancer. The ultimate performance of classification in this algorithm is also estimated as 0.8471, which, as compared to many other methods, has had a much better performance.

\subsubsection{Comparison with the state-of-the-art}

At the first stage of our algorithm, whose aim is to classify the mammography images into normal and abnormal categories, we did not attempt to remove the tags and also the pectoral muscle. This is because of the high ability of this network to be robust to noise, which contribute to learning the useful characteristics and ignore the useless ones. Other methods, such as Support Vector Machine (SVM), Naïve Bayes\cite{NaiveBayes10}, and mass learning (AdaBoost)\cite{AdaBost14}, eliminate the undesirable items in order to have an acceptable functionality.

Another item to be noted in respect to these networks is the automatic detection and extraction of the features; whereas, other algorithms are not able to extract the useful features on their own and this is something to be done manually. 

In order to make a comparison between our algorithm and other ones (\ref{table_2}), the features which are learned by CNN are being applied. These features are the learned features from the last convolutional layer, which are all located in the last CNN layer, meaning that the fully connected layers are located over of each other.

Thus, dimension of feature vector is reduced, using the Principle Component Analysis (PCA), and then these features are used to make comparison with other algorithms.  

In so doing, the number of feature maps taken from CNN equals 300. As mentioned before, to test the algorithms, the dimensions of the feature vector is reduced to 100 and 200, using the PCA, and the comparison is made with different dimensions of 100, 200, 300.

Here, we also use the Receiver Operating Characteristic curve (ROC) (a-c in Figure \ref{roc_fig}) to compare and evaluate our suggested method with comparing methods. Moreover, we would investigate the algorithm, using the accuracy, sensitivity, precision, recall, geometrical mean, and F-measure factors (Table \ref{table_2}).

As shown in Table \ref{table_2}, our suggested algorithm has a higher level of performance in comparison to the other algorithms. The performance can be estimated by considering the evaluation measures.

The precision of our method in correct classification of cases with cancer is 0.9288 and the accuracy of the diagnosis of cancerous and noncancerous cases is 0.7806. Among all the algorithms which are compared with ours, Naïve Baves classifier, has shown better results than other classifications methods. With 100, 200, and 300 number of features, Naïve Baves has estimated the precision of diagnosis of cancerous cases as 0.9021, 0.8961, and 0.8902 respectively; also, the accuracy of the diagnosis of cancerous and noncancerous cases equals 0.7495, 0.7553, and 0.7631 respectively.

\subsection{Segmentation Results}
In the second stage, our algorithm is responsible to segment the location of the tumor in the mammography images. This stage also consists of the two sub-stages: pre-processing and segmentation.  

\subsubsection{Pre-processing} 
The pre-processing stage consists of 3 sections: noise reduction, the elimination of artifacts, and pectoral muscle removal. See Fig. \ref{preprocess_fig}.

In order to reduce noise in the second stage, BM3D algorithm is used. After noise reduction, we attempt to remove the artifacts and tags from the mammography images. In this study the artifacts removal is also based on the thresholding. The method and the settings will be discussed in the following.  

To remove the radiopaque artifacts and tags, the median filter is primarily applied with size of 10*10. Then we used data normalization to avoid the information redundancy. In so doing we set the minimum and maximum brightness on 60 and 210 respectively. After that the grey level mammography images are transformed into binary images, using the Otso thresholding.

\subsubsection{Segmentation, using the level-set based on spatial fuzzy clustering (LS-SFC)}

The level-set method has several controlling parameters whose accurate setting would cause quite acceptable results. These parameters are estimated through the results of the fuzzy clustering. 

The Fig. \ref{qualitative_results} illustrates the results obtained from the level-set area segmentation algorithm based on spatial fuzzy clustering. In this algorithm the spatial fuzzy clustering estimates the tumor borders; then the results of this estimation are applied for the initialization of the level-set area segmentation. Consequently, the level-set algorithm would automatically estimate the controlling parameters from the fuzzy clustering and would initiate the evaluation from in the vicinity of the actual border. Ultimately, the level-set evolves in the approximation of this border; simply put, it improves the area segmentation obtained from the spatial fuzzy clustering. In this figure the primary and final area segmentation is done after 200 times repetition and, eventually, in last row of Fig. \ref{qualitative_results} the accurate location of the tumor is determined.

In Fig. \ref{qualitative_results}, first row shows the input images.

Second row shows the pre-processed images after denoising and pectoral removal.

Third row of Fig. \ref{qualitative_results}: 3D diagram obtained from the final valuing of the level-set based on spatial fuzzy clustering.

Last row of Fig. \ref{qualitative_results} revealing the tumor in image.

\section{Conclusion}
According to the statistics announced by the World Health Organization (WHO), breast cancer is the second most common type of cancer among women. Mammography is known as the most effective technique for early diagnosis of this type of cancer. It can detect the cancer 10 years prior to its appearance in breast. The algorithm which is offered by the current study is a two-level algorithm. In the first level, the mammography images are classified through the convolutional neural network (CNN).

In the second level, the tumor is revealed in the images. Having been classified into normal or abnormal categories, the images are used at this point for the tumor segmentation. At this stage, first, the images are being improved to increase accuracy. This improvement includes noise reduction by means of the block matching and 3D filtering, tag elimination, and pectoral muscle removal form the images. After this step, which is known as the pre-processing level, the location of the tumors is identified in the images. In so doing, the combined algorithm of the level-set based on spatial fuzzy clustering is applied.

This algorithm starts with a spatial fuzzy clustering and its results are being used for the initialization of the level-set area segmentation and, thus, the tumors are identified in the images.


\bibliographystyle{spmpsci}      

\end{document}